\documentclass[10pt,twocolumn,letterpaper]{article}

\usepackage[pagenumbers]{cvpr} 
\usepackage[title]{appendix}

\usepackage{graphicx}
\usepackage{amsmath}
\usepackage{amssymb}
\usepackage{booktabs}
\usepackage{xcolor}
\usepackage{pifont}
\usepackage{multirow}
\usepackage{algorithm2e}
\usepackage{nicefrac}
\usepackage{comment}
\usepackage{microtype}
\RestyleAlgo{ruled} 

\DeclareMathOperator*{\argmax}{arg\,max}

\usepackage[pagebackref,breaklinks,colorlinks]{hyperref}

\usepackage[capitalize]{cleveref}
\crefname{section}{Sec.}{Secs.}
\Crefname{section}{Section}{Sections}
\Crefname{table}{Table}{Tables}
\crefname{table}{Tab.}{Tabs.}

\begin{document}
\everypar{\looseness=-1}
\title{Unsupervised Adaptation of Semantic Segmentation Models without Source Data}

\author{
Sujoy Paul, Ansh Khurana, Gaurav Aggarwal \\
Google Research \\
\tt\small \{sujoyp, anshkhurana, gauravaggarwal\}@google.com}

\maketitle

\begin{abstract}
   We consider the novel problem of unsupervised domain adaptation of source models, without access to the source data for semantic segmentation. Unsupervised domain adaptation aims to adapt a model learned on the labeled source data, to a new unlabeled target dataset. Existing methods assume that the source data is available along with the target data during adaptation. However, in practical scenarios, we may only have access to the source model and the unlabeled target data, but not the labeled source, due to reasons such as privacy, storage, etc. In this work, we propose a self-training approach to extract the knowledge from the source model. To compensate for the distribution shift from source to target, we first update only the normalization parameters of the network with the unlabeled target data. Then we employ confidence-filtered pseudo labeling and enforce consistencies against certain transformations. Despite being very simple and intuitive, our framework is able to achieve significant performance gains compared to directly applying the source model on the target data as reflected in our extensive experiments and ablation studies. In fact, the performance is just a few points away from the recent state-of-the-art methods which use source data for adaptation. We further demonstrate the generalisability of the proposed approach for fully test-time adaptation setting, where we do not need any target training data and adapt only during test-time. 

\end{abstract}

\section{Introduction}

\begin{figure}[th]
    \includegraphics[scale=0.37]{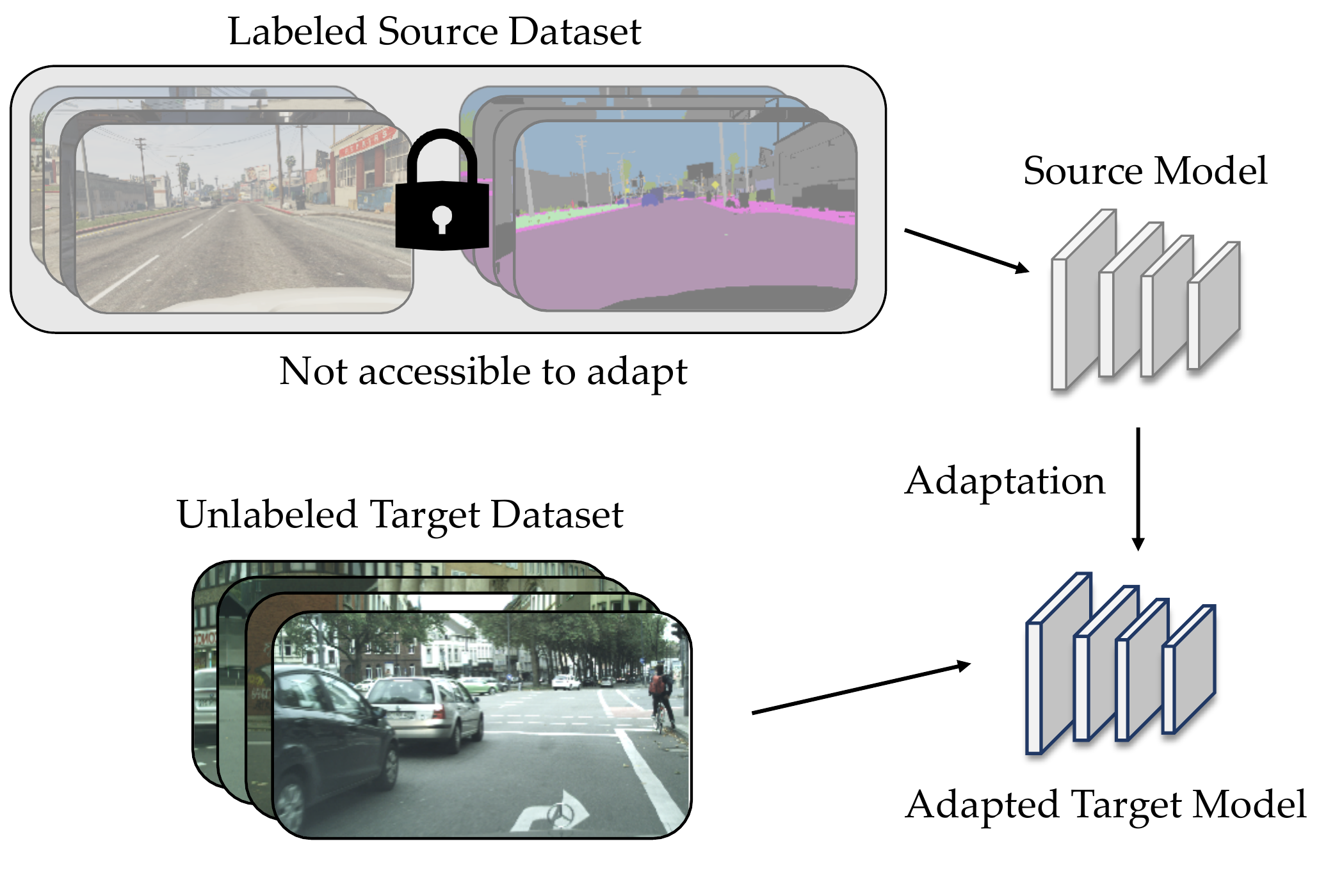}
    \caption{\textbf{Overview.} Unsupervised domain adaptation methods for semantic segmentation in literature use labeled source dataset while adapting to an unlabeled target dataset. However, in this work, we only use the knowledge from the source model for adaptation without access to the source data. 
    }
    \label{fig:teaser}
\end{figure}

Deep neural networks have shown immense success for dense prediction tasks, such as semantic segmentation~\cite{deeplab}. However, models learned on one dataset, say the source, may not generalize well to a new dataset, say the target. 
Annotating every new target dataset is expensive, especially for the semantic segmentation task. For example, a single image from the Cityscapes dataset required about 1.5 hours to annotate~\cite{cityscapes}.  To avoid the requirement of annotated target data, Unsupervised Domain Adaptation (UDA) methods~\cite{tzeng2017adversarial,Hoffman_ICML_2018} adapt the model learned with labeled source data to the target dataset, without needing annotated target images.

In the literature, UDA methods for semantic segmentation assume that data points from both the source and the target datasets are available for adaptation. However, such an assumption may not be practical, where either the source dataset cannot be shared to maintain privacy or the source dataset may take up a lot of storage space. In this case, one has access to only the light-weight source model which {\textit{was}} trained using labeled source data. This would restrict sharing private information as well as require much less storage. For example, storing the GTA5 dataset~\cite{Richter_ECCV_2016} used as source in most UDA methods, needs $57$ GB of storage whereas a model trained using the same data requires only $0.17$ GB.  With this motivation, in this work, we do not assume access to any source data and adapt the source model using only the unlabeled target data for semantic segmentation. Figure~\ref{fig:teaser} illustrates the problem pictorially.

Before diving into the details of source-free adaptation, let's study the techniques used in UDA for semantic segmentation using source data. They can be categorized mainly into three groups: 1) output space adaptation~\cite{tsai2018learning,Hoffman_CoRR_2016} that aligns the source and target output distributions, 2) pixel space adaptation~\cite{Hoffman_ICML_2018,Choi_ICCV19} that translates the source image distribution to the target distribution, and uses them for training, and finally, 3) pseudo labeling~\cite{Zou_ECCV_2018,Lian_ICCV19} that predicts the labels of the target images using the source model and uses them for training. When we do not have access to the source data, the first two methods are not even applicable. 
Recent methods on UDA for classification task \cite{liang2020we,yeh2021sofa} without source data largely follow the third strategy, i.e., pseudo-labeling. A few methods in this line of work also generate data from the target distribution to boost the performance \cite{li2020model}. Very recently published works \cite{Liu_CVPR_2021, S_2021_CVPR} for source-free adaptation of segmentation models are no different. Additionally, \cite{Liu_CVPR_2021} also follow the first option by dividing the target data into easy (source) and hard (target) group and train an additional discriminator. 

In this work, we propose a novel consistency based self-training approach to adapt the source model using unlabeled target images. As the target data comes from a shifted distribution than the source, we first update the norm parameters of the network using the unlabeled target data. Then, we extract knowledge from the source model by pseudo-labeling the unlabeled target. The extracted pseudo-labels should be consistent against certain spatial transformations on images such as rotation, flipping, mirroring, and image augmentations such as cutout, Gaussian blur, etc. We use these consistencies to enhance the knowledge extracted from the source model. This can be posed as a constrained optimization problem, where in each iteration, we first compute the gradients of the target model using the actual pseudo-labels with respect to the source model, and then obtain the gradients using the consistency losses computed with respect to the current target model. We show via empirical analysis that these simple consistencies with pseudo-labeling help to considerably improve the performance of the model on the target images. We also create a pseudo image distribution by collaging pairs of images, which enhances the variety of images in the dataset, and helps to regularize the model.

We evaluate our framework on three different settings: GTA5 $\rightarrow$ Cityscapes and SYNTHIA $\rightarrow$ Cityscapes for outdoor scenes and the novel setting of SceneNet $\rightarrow$ SUN for indoor scenes, showcasing the generic nature of our algorithm. There is no known effort in the literature to show domain adaptation results on the mentioned indoor setting. 
Further, to demonstrate the generalisability of our consistency based approach, we apply it in the fully test-time adaptation setting~\cite{wang2021tent}, where it outperforms loss functions proposed by recent methods dedicated for fully test time adaptation~\cite{mummadi2021testtime, wang2021tent} on all the above three dataset settings.
The \textbf{main contributions} of our work are:
\begin{itemize}
    \item We tackle the interesting problem of source-free adaptation for semantic segmentation and propose a self-training approach by imposing certain input-output consistencies to extract the knowledge from the source model. 
    \item Empirical analysis shows that our method establishes the new state-of-the-art for source-free adaptation, and is only a few points away from the state-of-the-art which uses the entire labeled source dataset to adapt.
    \item Our method works well even for test time adaptation, where we do not access even the unlabeled target, and the adaptation is done individually for each test sample. 
\end{itemize}

\section{Related Works} \label{sec:related_works}
\begin{figure*}[!t]
    \centering
    \includegraphics[scale=0.38]{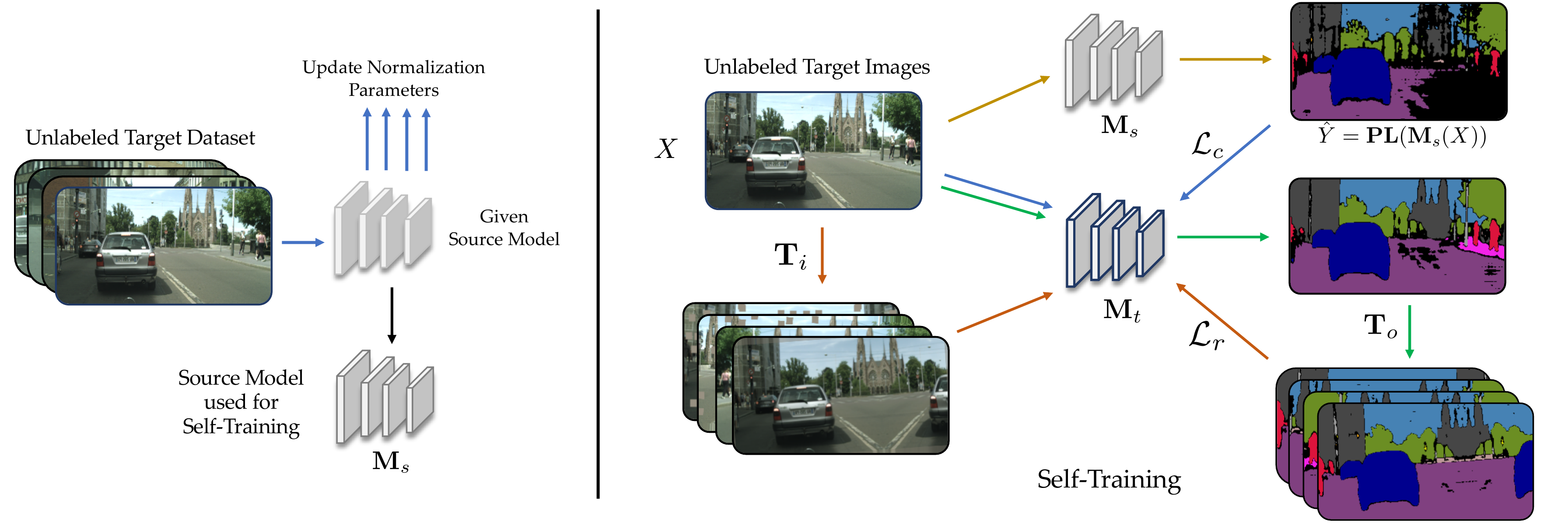}
    \caption{\textbf{Framework overview.} (Left) Given the source model, we first update the norm parameters of the network by just forward propagating the unlabeled target images, to obtain the model $\boldsymbol{\mathrm{M}}_s$. (Right) Then we learn a new target model $\boldsymbol{\mathrm{M}}_t$ using self-training. We first obtain the pseudo-labels using $\boldsymbol{\mathrm{M}}_s$ and compute the classification loss $\mathcal{L}_c$. Then, we transform the input image ($\boldsymbol{\mathrm{T}}_i$) as well as the target model's prediction ($\boldsymbol{\mathrm{T}}_o$) in the current iteration and impose certain consistencies via the loss function $\mathcal{L}_r$. On the right-hand side, each arrow color denotes one full pass, either forward or forward+backward propagation.}
    \label{fig:framework}
\end{figure*}

We divide the related works discussion in two parts: 1) methods using the source data to adapt (most works in literature), and 2) methods that do not use any source data, (a few recent efforts, mostly for classification and a couple for the segmentation task). 

{\flushleft \bf{UDA with Source Data.}}
UDA methods for the image classification task aim to align the source and target distributions. To do so, maximum mean discrepancy~\cite{long2015learning} and adversarial learning~\cite{ganin2016domain,tzeng2017adversarial} based approaches have been proposed. A few algorithms focus on improving deep models \cite{long2016unsupervised,Saito_CVPR_2018,Lee_CVPR_2019,dai2019adaptation} and translating the input source images to the target domain \cite{Bousmalis_CVPR_2017,Hoffman_ICML_2018,Luan_CVPR_2019}. 
For the more challenging task of semantic segmentation, existing UDA methods can be categorized primarily into three groups: output alignment, pixel-adaptation and pseudo-labeling. In output alignment, the methods aim at aligning the output or feature distribution between the source and target domains\cite{tsai2018learning,Chen_CVPR_2018, Chen_ICCV_2017,Hoffman_CoRR_2016,Zhang_ICCV_2017, araslanov2021self}. 
For pixel alignment, similar to as in image classification, the algorithms translate the input images to the target domain by changing style, while preserving content information \cite{Chang_CVPR_2019,Choi_ICCV19,Hoffman_ICML_2018,Murez_CVPR_2018,Wu_ECCV_2018,Zhang_CVPR_2018,yang2020fda}. 
For pseudo-labeling, methods aim to generate pixel-wise pseudo labels on the target images, which is utilized to fine-tune the segmentation model trained on the source~\cite{Saleh_ECCV_2018,Zou_ECCV_2018,Lian_ICCV19,zhang2019category,li2020content, pan2020unsupervised,shin2020two,mei2020instance,dong2020cscl}. 
There are also several methods in literature that try to combine these strategies \cite{Du_ICCV19,Li_CVPR_2019,Tsai_DA4Seg_ICCV19,Vu_CVPR_2019,paul2020domain,wang2020differential,musto2020semantically,kim2020learning,lv2020cross,huang2020contextual,subhani2020learning}. Compared to these approaches, we assume no access to the source dataset, which is a more realistic setting but makes the task much more challenging.

{\flushleft \bf{UDA without Source Data.}}
Unlike the above methods, there have been a few works for classification tasks which do not use source data, but only the source model for adaptation. The methods involve - entropy minimization with divergence maximization \cite{liang2020we}, pseudo-labeling with self-reconstruction \cite{yeh2021sofa}, generating additional target images \cite{li2020model} and self-supervision \cite{xia2021adaptive}. 
For source-free adaptation for semantic segmentation, recently, \cite{Liu_CVPR_2021} proposed an algorithm that combines ideas from the above methods like image generation, pseudo-labeling and output space adaptation by dividing the target into easy and hard, and learning an additional discriminator to align them.  In \cite{S_2021_CVPR}, pseudeo-labeling and robustness to dropout randomness is used to tackle the problem, which performs very similar to our strong baseline of Pseudo-Label (PL). Pseudo-Labeling is also used in \cite{you2021domain} with the concept of negative learning, i.e., the pixels where the model is not confident towards a certain positive label, it may be confident towards multiple negative labels. Very recently, a method for robust training of source models is proposed in \cite{kundu2021generalize}, followed by pseudo-labeling on the target. In comparison to this method, we do not assume access to the source data or training process but only the trained source model. 
\section{Domain Adaptation from Source Models}

\begin{figure*}[!htpb]
    \centering
    \includegraphics[scale=0.4]{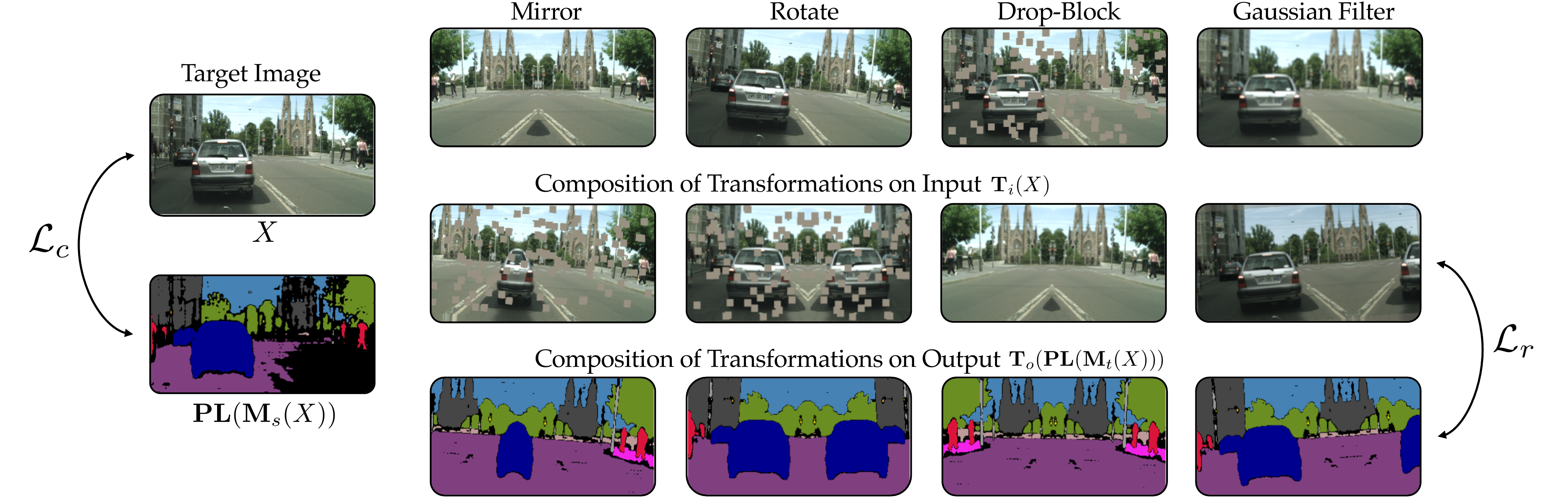}
    \vspace{1mm}
    \caption{\textbf{Transforms.} This figure shows the different transforms used in our algorithm. The left-most side shows an image with the pseudo-labels obtained from the source model $\boldsymbol{\mathrm{M}}_s$, which are used to compute the loss $\mathcal{L}_c$. The top row shows the transformations individually, followed by their random compositions in the second row, and the last row shows the transformations on the output of the target model for a certain iteration, which are used a ground-truth for computing loss $\mathcal{L}_r$.}
    \label{fig:transforms}
\end{figure*}

In this section, we formally define the problem statement, followed by an overview of our framework, and finally a detailed description of each blocks in our framework.

\subsection{Problem Statement}
Consider that we have a source model for semantic segmentation, which needs to be adapted to a new target domain, given only unlabeled target images, without access to the source images on which the source model was trained. Formally, consider we have a source model $\boldsymbol{\mathrm{M_s}}: \mathbb{R}^{H \times W \times 3} \rightarrow \mathbb{R}^{H \times W \times C}$, which takes as input an RGB image and predicts for every pixel its category. $H, W$ are the height and width of the image and $C$ is the number of categories/labels. Now, given an unlabeled target dataset, $\mathcal{D}_t=\{X_i\}_{i=1}^n$, our goal is to adapt the source model $\boldsymbol{\mathrm{M_s}}$ to a new model $\boldsymbol{\mathrm{M_t}}$, such that it performs better on images drawn from the target distribution, than directly using the source model on the target images. 

\subsection{Overall Framework} 
An overview of our proposed framework is presented in Figure~\ref{fig:framework}. As the target images come from a different distribution than what the source model is trained with, we first update only the norm parameters of the network, i.e., mean and variance of BatchNorm or InstanceNorm, by just forward propagating the target domain images $\mathcal{D}_t$ through the given source model. This does not change the network weights, but only the norm parameters are updated. After that, we use the updated source model $\boldsymbol{\mathrm{M}}_s$ to pseudo-label the target images only for the confident pixels. Along with these pseudo-labels, we impose two different types of consistency constraints using augment and spatial transformations. 
This helps to efficiently extract knowledge from the source model that in turn helps the adapted model perform better on the target images. Additionally, we create a pseudo image distribution by collaging pairs of images, thus creating a wider variety of image compared to the target dataset. We learn a new target model $\boldsymbol{\mathrm{M}}_t$ by simultaneously optimizing for the pseudo-labels with the consistency constraints.

\subsection{Update Normalization Parameters} \label{sec:update_norm}
BatchNorm \cite{Ioffe_ICML_2015} and InstanceNorm \cite{ulyanov2016instance} are commonly used in deep neural networks to avoid over-fitting, stabilize training, and faster convergence. For every activation layer in the network, these normalization methods estimate two parameters during training, viz., the mean and variance, which are then used in the testing phase. However, these parameters are only a good estimate for the images in the training distribution. Thus, in domain adaptation, where the target data may belong to a different distribution than the source, it may be useful to first update the mean and variance of the normalization layers, using the unlabeled target data. This is inspired by works in literature~\cite{li2016revisiting,chang2019domain} where they estimate domain-specific normalization parameters, while having access to both the source and the target data, during training. 

In order to update the normalization statistics, we only forward propagate the unlabeled target images through the network for one epoch and use the following update rule for a particular layer in each iteration.
\begin{align}
    \mu &\leftarrow (1-\nicefrac{1}{i}) \mu + \frac{1}{i} \mu_i \\
    \sigma^2 &\leftarrow (1-\nicefrac{1}{i}) \sigma^2 + \frac{1}{i} \sigma^2_i
    \label{eq:bn_update}
\end{align}
where $i = \{1, \dots, n\}$. $\mu_i$ and $\sigma^2_i$ are the mean and variance of that layer with $X_i$ as input. Note that we use instance normalization in all our experiments. This is because in semantic segmentation, to handle high resolution images, the batch size can typically only be one. Further, to estimate the mean and variance on the new target data, we need to forward propagate each target image only once, without any backward propagation. This step only updates the normalization parameters, mean and variance, and not the weights of the network. We call this updated model $\boldsymbol{\mathrm{M}}_s$, which we use to learn the target model $\boldsymbol{\mathrm{M}}_t$, as discussed next.

\subsection{Pseudo-Labeling with Constraints}

Given the normalization updated source model $\boldsymbol{\mathrm{M}}_s$, we use it to label the target images via confidence-filtered pseudo labeling. Additionally, we impose consistency using certain transforms, which gives the model an opportunity to improve beyond learning from just the pseudo-labeled dataset.

\subsubsection{Pseudo-Labeling:}
The pseudo-labeling function $\hat{Y}=\boldsymbol{\mathrm{PL}}(\boldsymbol{\mathrm{M}}(X))$ to label an image $X$ with a model $\boldsymbol{\mathrm{M}}$ can be defined as follows:
\begin{equation}
    \hat{Y}^{x,y} =
    \begin{cases}
      j^* = \argmax_j [\boldsymbol{\mathrm{M}}(X)]^{x,y}, & {\scriptstyle \text{if}  \max_j [\boldsymbol{\mathrm{M}}(X)]^{x,y} > p_{j^*}}\\
      \text{No Label}, & \text{otherwise}
    \end{cases}
    \label{eqn:pseudolabel}
\end{equation}
where the superscript $x,y$, represents the spatial co-ordinates, $j \in \{1, \dots, C\}$, and $[\boldsymbol{\mathrm{M}}(X)]^{x,y}$ is a $C$-dimensional vector summing to 1, representing the pixel's prediction. $p_j$ is the threshold for prediction confidence of label $j$, 
which will be discussed in more detail subsequently.
We label the unlabeled target dataset $\mathcal{D}_t$ using the pseudo-labeling function to obtain a new dataset $\hat{\mathcal{D}}_t=\{X_i, \hat{Y}_i\}_{i=1}^n$, where $\hat{Y}_i=\boldsymbol{\mathrm{PL}}(\boldsymbol{\mathrm{M}}_s(X_i))$.

We can train a new model $\boldsymbol{\mathrm{M}}_t$ using the pseudo-labeled set $\hat{\mathcal{D}}_t$, which would reduce the uncertainties in the source model $\boldsymbol{\mathrm{M}}_s$, offering a better hypothesis on the target. 
However, the model can only be as good as the pseudo-labels, without much room for improvement. We observe that certain transforms on the target input actually lead to undesirable changes in the output. Thus, we constrain these to desirable outputs, which acts as a signal to improve beyond the pseudo-labeled set.

\subsubsection{Consistency as Constraint} The target network learned using the pseudo labels from Eqn.~\ref{eqn:pseudolabel} can only be as good as the pseudo-labels and it may overfit to that particular set $\hat{\mathcal{D}}_t$.  Because of this, we observe that under certain transformations to the input image such as cutout, Gaussian blur, image rotation, image mirroring, etc., the network does not predict consistent labels. By consistent, we mean labels similar to what one would obtain without applying these transformations. This observation motivates us to use this consistency as constraint while learning from the pseudo-labeled dataset $\hat{\mathcal{D}}_t$, which acts as an additional signal to improve beyond pseudo-labeling. We use two different types of image transformations namely - augment and spatial transforms, as discussed next. 
Moreover, we create additional images (collages)
by joining pairs of images and apply similar consistency constraints to further enhance the model.
{\flushleft \bf{Augment Transforms $(\mathcal{T}_a)$}:}
Cutout and Gaussian blur belong to this transform, as shown in Figure~\ref{fig:transforms}. In this case, we constrain the output to be similar even after applying these transforms, i.e., if we consider a transform $\boldsymbol{\mathrm{T}} \in \mathcal{T}_a$, then the consistency constraint can be expressed as
\vspace{-1mm}
\begin{equation}
    \boldsymbol{\mathrm{M}}_t(\boldsymbol{\mathrm{T}}(X)) \approx \boldsymbol{\mathrm{M}}_t(X)
\end{equation}
In cutout, we randomly choose $k$ blocks of size $b \times b$ from the image, and set them to $0$, where $\frac{kb^2}{HW} \approx p$, with two free parameters $b, p$. $H$ and $W$ are the image height and width. For Gaussian blur, we choose a kernel size and filter variance. Please refer to the appendix for details on the parameter values used for these transformations (Figure~\ref{fig:cutout_ablation}).

{\flushleft \bf{Spatial Transforms $(\mathcal{T}_s)$}:}
Image mirroring and rotation fall under this transform set as shown in Figure~\ref{fig:transforms}. Compared to augmentations, in this case, the output should not be the same as obtained from the original image, but should also be transformed. Formally, considering a transformation $\boldsymbol{\mathrm{T}} \in \mathcal{T}_s$, the consistency constraint can be expressed as
\vspace{-1mm}
\begin{equation}
\boldsymbol{\mathrm{M}}_t(\boldsymbol{\mathrm{T}}(X)) \approx \boldsymbol{\mathrm{T}}(\boldsymbol{\mathrm{M}}_t(X))    
\end{equation} 
Note that the order of the model and transform is switched in this case. For image mirroring, we first randomly choose a column, vertically splitting the image, and mirror the larger side of the image onto the smaller side. For rotation, we randomly choose the rotation degree between $[-5^{\circ}, +5^{\circ}]$. 

\begin{figure}
    \centering
    \includegraphics[scale=0.28]{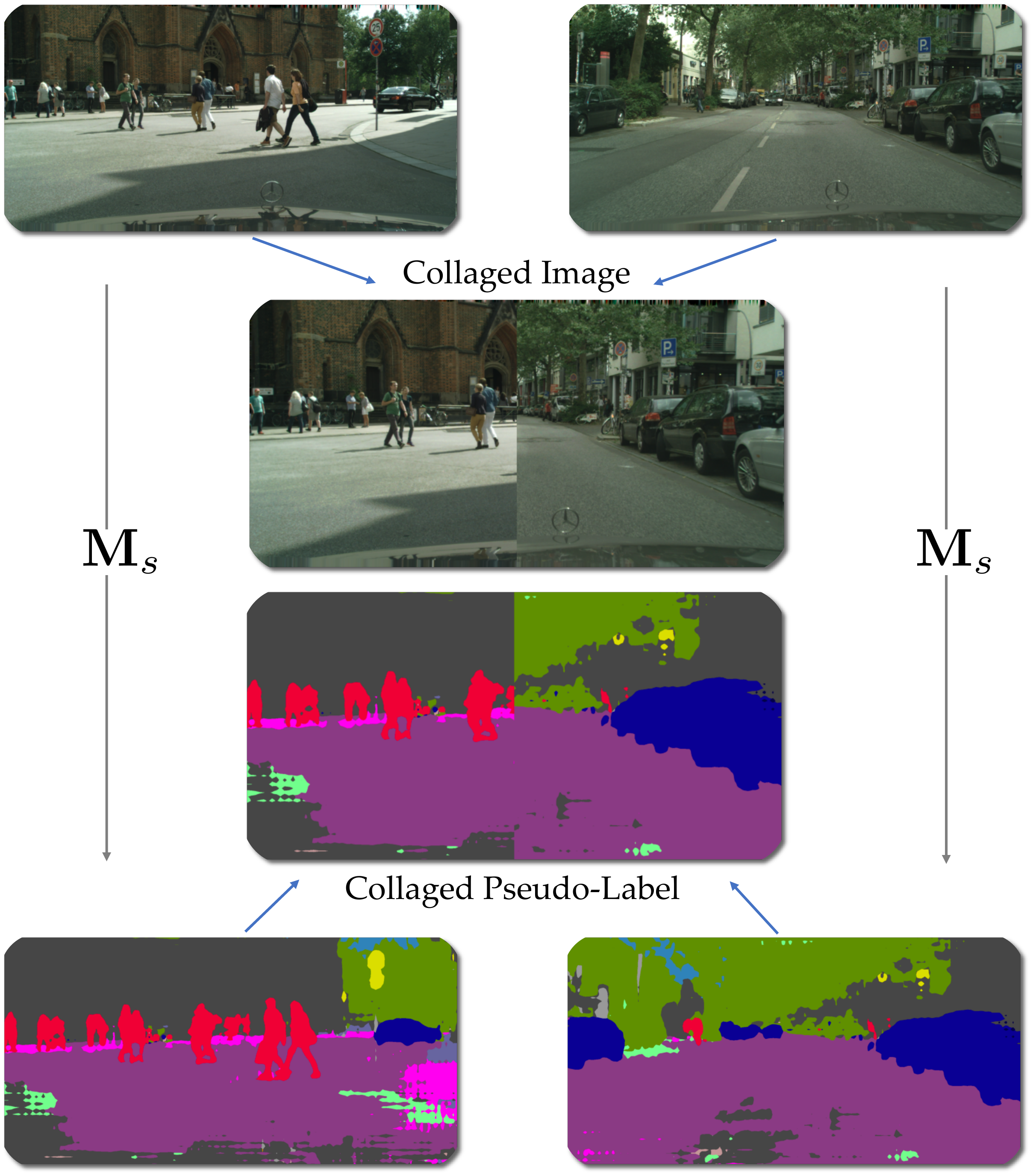}
    \caption{Process of image and pseudo-label collage creation. We use such collages to learn the target model.}
    \label{fig:pairs}
\end{figure}

{\flushleft \bf{Learning from Image Collages.}} Mixup \cite{zhang2017mixup} uses weighted combination of image pairs and their labels to regularize model training. Motivated by this, we propose to combine two images into a single image, but in the spatial domain, as shown in Figure \ref{fig:pairs}. Given a pair of images $X_i, X_j$, the collage image is constructed by concatenating the first half of $X_i$ with the second half of $X_j$ along the width. Now, to obtain the pseudo-labels for the collage image, we also concatenate the pseudo-labels $\boldsymbol{\mathrm{PL}}(\boldsymbol{\mathrm{M}}_s(X_i))$ and $\boldsymbol{\mathrm{PL}}(\boldsymbol{\mathrm{M}}_s(X_j))$, as shown in Figure \ref{fig:pairs}. 
Note that in our algorithm, we learn only from these collage images, and $X$ from here onwards denotes these collage images. This creates a wider variety of images that helps to improve the target model. 

{\flushleft \bf{Optimization Problem}:} Using these consistency constraints, we solve the following optimization problem: 
\begin{align}
    \min_{\boldsymbol{\mathrm{M}_t}} & \quad \mathcal{L}_c = \sum_{X \in \mathcal{D}_t } l_c\Big(\boldsymbol{\mathrm{M}}_t(X), \boldsymbol{\mathrm{PL}}(\boldsymbol{\mathrm{M}}_s(X))\Big) \label{eq:pl_loss} \\
    \text{s.t.} & \quad  \boldsymbol{\mathrm{M}}_t(\boldsymbol{\mathrm{T}}(X)) \approx \boldsymbol{\mathrm{M}}_t(X) \; \; \; \; \; \; \; \; \forall X \in \mathcal{D}_t, \boldsymbol{\mathrm{T}} \in \mathcal{T}_a \label{eq:augment_transform} \\
    & \quad  \boldsymbol{\mathrm{M}}_t(\boldsymbol{\mathrm{T}}(X)) \approx \boldsymbol{\mathrm{T}}(\boldsymbol{\mathrm{M}}_t(X)) \; \; \; \forall X \in \mathcal{D}_t, \boldsymbol{\mathrm{T}} \in \mathcal{T}_s \label{eq:spatial_transform}
\end{align}
The objective function is thus to optimize for the pseudo-labels for the confident pixels as obtained from Eqn. \ref{eqn:pseudolabel} using cross-entropy loss $l_c$, such that the output of the target model is consistent with augment and spatial transforms. In practice, we compose the two sets of transforms to strengthen its power. To do so, in each iteration, we randomly choose a set of transformations $\mathcal{T} = \{\boldsymbol{\mathrm{T}}_0, \dots, \boldsymbol{\mathrm{T}}_k\} \in \mathcal{P}(\mathcal{T}_a \cup \mathcal{T}_s) \setminus \emptyset$. Due to the differences in the consistencies, the transforms need to be composed differently for input and output. The input transform is a composition of all the transforms, $\boldsymbol{\mathrm{T}}_i = \boldsymbol{\mathrm{T}}_0 \circ \dots \circ \boldsymbol{\mathrm{T}}_k$. But to compose the output transform, we remove those drawn from $\mathcal{T}_a$, and then compose the rest in the same manner as input transform to obtain $\boldsymbol{\mathrm{T}}_o$. Then, the constraints can be simplified as $\boldsymbol{\mathrm{M}}_t(\boldsymbol{\mathrm{T}}_i(X)) \approx \boldsymbol{\mathrm{T}}_o(\boldsymbol{\mathrm{M}}_t(X))$.

We approximate the constraint using two losses, and impose it via the penalty method. For every iteration, we consider $\boldsymbol{\mathrm{T}}_o(\boldsymbol{\mathrm{M}}_t(X))$ i.e., the transformed outputs from the current iteration target model, as ground-truths for the losses, and do not pass gradients through it, as otherwise trivial solutions can satisfy the constraints. The two losses are:
\begin{align}
    \mathcal{L}_{r} &= \sum_{X \in \mathcal{D}_t} l_c\Big(\boldsymbol{\mathrm{M}}_t(\boldsymbol{\mathrm{T}}_i(X)), \boldsymbol{\mathrm{T}}_o(\boldsymbol{\mathrm{M}}_t(X)) \Big)  \nonumber \\ 
    & \quad \quad \quad \quad + l_c\Big(\boldsymbol{\mathrm{M}}_t(\boldsymbol{\mathrm{T}}_i(X)), \boldsymbol{\mathrm{T}}_o(\boldsymbol{\mathrm{PL}}(\boldsymbol{\mathrm{M}}_t(X)))\Big)
    \label{eqn:constrain_loss}
\end{align}
where $l_c$ is the cross-entropy loss. The first loss is a form of soft pseudo-labeling and computed for every pixel, whereas the second loss is hard pseudo-labeling, which only computes the loss for the confident pixels.

\vspace{2mm}
\noindent
\textbf{Parameter Updates.} We optimize using Stochastic Gradient Descent. Note that we use the pseudo-labels using $\boldsymbol{\mathrm{M}}_s$ for ground-truth in the loss function of Eqn. \ref{eq:pl_loss} and $\boldsymbol{\mathrm{M}}_t$ in the current iteration as the ground-truth for the loss function in Eqn. \ref{eqn:constrain_loss}. For clarity, the SGD updates for the $k^{th}$ iteration, with $\eta$ as the learning rate can be expressed as follows:
\begin{equation}
    \boldsymbol{\mathrm{M}}_t^{k+1} = \boldsymbol{\mathrm{M}}_t^{k} - \eta \nabla_{\boldsymbol{\mathrm{M}}_t} \Big(\mathcal{L}_c(\boldsymbol{\mathrm{M}}_s) + \mathcal{L}_r(\boldsymbol{\mathrm{M}}_t^k) \Big)
\end{equation}
where the argument in the brackets represent the models used to obtain the ground truths to compute the losses.

\noindent
{\flushleft \bf{Pseudo-label Thresholds.}}
To obtain the pseudo-labels using the source model $\mathrm{M}_s$, we set the thresholds to be $\min(0.9, \text{median of label-wise confidence})$, where the median is computed over all pixel predictions of the dataset for that label. The confidence distribution varies across labels, which motivates us to choose a label-dependent threshold rather than a fixed threshold for all labels. This strategy is commonly used in literature, and we also observe a boost in performance using this strategy. Now, as the target model $\boldsymbol{\mathrm{M}}_t$ evolves over time, we do not keep the same threshold to obtain the pseudo-label for its predictions, but rather update it using exponential moving average as follows:
\begin{equation}
    \boldsymbol{\mathrm{p}} = \lambda \boldsymbol{\mathrm{p}} + (1-\lambda) \boldsymbol{\mathrm{p}}_k 
    \label{eqn:threshold_ema}
\end{equation}
where $\boldsymbol{\mathrm{p}}$ is the threshold vector over the $C$ categories and used for the target model, $\boldsymbol{\mathrm{p}}_k$ is the threshold vector obtained for the image in the $k^{th}$ iteration using the median strategy mentioned above, computed over all pixel predictions of that image for that label.

\section{Experiments}
\begin{table} [t]
	\caption{\small Results of adapting GTA5 to Cityscapes.
	The top group are methods which use source data during adaptation, while the bottom two row groups do not use any source data to adapt. Note that the middle row group use DeepLabv3+ResNet-50 as their backbone while all other methods use DeepLabv2+ResNet-101.
	}
	\label{table:gta5_cityscapes}
	\scriptsize
	\centering
	\renewcommand{\arraystretch}{1.2}
	\resizebox{0.48\textwidth}{!}{
	\begin{tabular}{llccc}
		\toprule
		
		
		Source & Method &Stuff&Things& mIoU\\
		\midrule
        \multirow{7}{*}{Yes}& AdaptOutput~\cite{tsai2018learning} &52.2&33.6& 41.4 \\
                            & AdvEnt~\cite{Vu_CVPR_2019} &57.0&37.1& 45.5 \\
                            & SSF-DAN~\cite{Du_ICCV19} &57.4&36.7& 45.4 \\
                            & BDL~\cite{Li_CVPR_2019} &61.7&38.9& 48.5 \\
                            & CAG \cite{zhang2019category} &61.2&42.0& 50.2 \\
                            & WeakDA \cite{paul2020domain} &61.0&38.8& 48.2 \\
                            & Stuff \cite{wang2020differential} &62.1&39.9& 49.2 \\
                            & FDA \cite{yang2020fda} &60.3&43.2& 50.4 \\
                            & SAC \cite{araslanov2021self} & 64.3 & 46.1 & 53.8 \\
        \midrule
        {No} & SFDA \cite{Liu_CVPR_2021} &56.7&33.3& 43.2 \\
             
        \midrule
		\multirow{3}{*}{No} & No Adapt. &45.6&31.6& 37.5 \\
		                    & URMA \cite{S_2021_CVPR} &60.3&34.0& 45.1 \\
		                    & LD\cite{you2021domain} &60.1&34.9& 45.5 \\
                            & Ours & 59.0 & 41.3 & \textbf{48.8}  \\
		\bottomrule
	\end{tabular}
	}
\end{table}

\begin{table} [t]
	\caption{ \small
		Results of adapting SYNTHIA to Cityscapes.
		The top group are methods which use source data to adapt, while the bottom two row groups do not use source data to adapt. mIoU and mIoU$^\ast$ are averaged over 16 and 13 categories. Note that the middle row group use DeepLabv3+ResNet-50 as the backbone while all other methods use DeepLabv2+ResNet-101. Stuff and Things mIoU is computed over the 13 categories for which all works report.
	}
	\label{table:synthia_cityscapes}
	\scriptsize
	\centering
	\renewcommand{\arraystretch}{1.3}
	\resizebox{0.48\textwidth}{!}{
	\begin{tabular}{llcccc}
		\toprule
		
		Source & Method & Stuff & Things & mIoU & mIoU$^\ast$ \\
		\midrule
		\multirow{7}{*}{Yes}& AdaptOutput~\cite{tsai2018learning} &70.8&30.4& 39.5 & 45.9 \\
                            & AdvEnt~\cite{Vu_CVPR_2019} &74.4&31.5& 41.2 & 48.0 \\
                            & SSF-DAN~\cite{Du_ICCV19} &74.6&34.6& - & 50.0 \\
                            & CAG \cite{zhang2019category} &73.9&37.4& 44.5 & 51.4 \\
                            & WeakDA \cite{paul2020domain} &78.5&35.3& 44.3 & 51.9 \\
                            & Stuff \cite{wang2020differential} &73.9&38.5& - & 52.1  \\
                            & FDA \cite{yang2020fda} &66.4&43.8& - & 52.5 \\
                            & SAC \cite{araslanov2021self} & 79.7 & 46.5 & 52.6 & 59.3 \\
		\midrule
        {No} & SFDA \cite{Liu_CVPR_2021} &76.8&26.5& 39.2 & 45.9 \\
        \midrule
		\multirow{3}{*}{No} & No Adapt. &57.0&23.9& 32.1 & 36.7  \\
		                    & URMA \cite{S_2021_CVPR} &64.9&32.6& 39.6 & 45.0 \\
		                    & LD\cite{you2021domain} &70.0&37.7& 42.6 & 50.1 \\
                            & Ours & 69.7 & 37.9 & \textbf{43.7} & \textbf{50.2} \\
		\bottomrule
	\end{tabular}
	}
\end{table}

{\flushleft \bf{Datasets:}}
We evaluate our framework on three combinations of source $\rightarrow$ target datasets covering both outdoor and indoor scenes. For outdoor, we evaluate on GTA5 \cite{Richter_ECCV_2016} $\rightarrow$ Cityscapes \cite{cityscapes} and SYNTHIA \cite{Ros_CVPR_2016} $\rightarrow$ Cityscapes. For indoor, we use SceneNet \cite{mccormac2017scenenet} $\rightarrow$ SUN \cite{song2015sun}.  Note that unlike existing literature, this work is the first to show results on the indoor setting, which depicts the generalization ability of our framework. The outdoor scene dataset has $19$ categories, whereas the indoor has $13$ categories. Please refer to Appendix~\ref{app:datasets} for more details.

{\flushleft \bf{Implementation Details:}}
To have a fair comparison with the works in literature, we use the Deeplab-V2 \cite{deeplab} with ResNet-101 \cite{He_CVPR_2016} as the network. We use one GPU to train our models with a batch size of $1$ in all experiments. We use SGD with an initial learning rate of $2.5\times 10^{-4}$ with polynomial decay of power $0.9$~\cite{deeplab}. We use the standard metric of mean intersection over union (mIoU)~\cite{deeplab} to evaluate all algorithms.

\begin{table*}[!htb]

    \begin{minipage}{.67\linewidth}
      	\caption{ 
		Results of adapting SceneNet to SUN. The top row group does not use any source data to adapt, while the bottom row uses full supervision on the target images.
	}
    \label{table:scenenet_sun}
    \renewcommand{\arraystretch}{1.1}
    \scriptsize
    \resizebox{1.0\linewidth}{!}{
	\begin{tabular}{lcccccccccccccc}
		\toprule
		
		
		Method & \rotatebox{70}{bed} & \rotatebox{70}{books} & \rotatebox{70}{ceiling} & \rotatebox{70}{chair} & \rotatebox{70}{floor} & \rotatebox{70}{furniture} & \rotatebox{70}{objects} & \rotatebox{70}{picture} & \rotatebox{70}{sofa} & \rotatebox{70}{table} & \rotatebox{70}{tv} & \rotatebox{70}{wall} & \rotatebox{70}{window} & mIoU \\
		
		\midrule
		No Adapt. & 19.6 & 10.1 & 22.2 & 42.5 & 65.4 & 21.3 & 13.4 & 20.9 & 18.2 & 27.1 & 6.2 & 57.1 & 20.2 & 26.5 \\
		Ours &  35.3 & 23.5 & 34.1 & 48.7 & 73.6 & 26.7 & 11.1 & 29.9 & 36.9 & 38.1 & 15.0 & 63.2 & 27.2 & \textbf{35.6} \\
		
		\midrule
		
		Full Supervision &  53.1 &  32.6 & 54.0 & 60.0 & 82.4 & 35.1 & 33.4 & 43.2 & 45.7 & 52.7 & 36.8 & 72.0 & 46.8 & 49.8 \\
		\bottomrule
	\end{tabular}
	}
    \end{minipage}%
    \hfill
    \begin{minipage}{.32\linewidth}
			\caption{Ablation of the transformations on GTA5 $\rightarrow$ Cityscapes. 
    	    }
			\label{table:transform_ablation}
            \renewcommand{\arraystretch}{1.1}
	        \scriptsize
			\resizebox{\linewidth}{!}{
			\begin{tabular}{cccccc}
		    \toprule
		    Collage & Mirror & Rotate & Gaussian & Cutout & mIoU \\
		    \midrule
		    \checkmark & & & & & 46.4 \\
		    & \checkmark & & & & 45.9 \\
		    & & \checkmark & & & 45.5 \\
		    & & & \checkmark & & 46.4 \\
            & & & & \checkmark & 45.8 \\
            \checkmark & \checkmark & \checkmark & \checkmark & \checkmark &  \textbf{48.8}\\
		    \bottomrule
	    \end{tabular}
	    }
    \end{minipage} 
\end{table*}

{\flushleft \bf{Comparison with state-of-the-art:}} 
We first compare our method with state-of-the-art in literature in Table~\ref{table:gta5_cityscapes} for GTA5 $\rightarrow$ Cityscapes, in Table~\ref{table:synthia_cityscapes} for SYNTHIA $\rightarrow$ Cityscapes, and SceneNet $\rightarrow$ SUN in Table~\ref{table:scenenet_sun}. Due to limited space, instead of presenting the category-wise performances, we club them into COCO-style Stuff and Things in Table \ref{table:gta5_cityscapes} and \ref{table:synthia_cityscapes}, and present the category-wise results in appendix (Table~\ref{table:gta5_cityscapes} and Table~\ref{table:synthia_cityscapes}). Stuff includes road, sidewalk, building, wall, fence, veg, terrain and sky; while Things include sign, person, rider, car, truck, bus, train, mbike, bike, light and pole. ``No Adapt." means applying the source model directly on the target data, without adaptation. Also, as this is the first work which presents domain adaptation results on SceneNet $\rightarrow$ SUN, we include the result with full-supervision on the target, in Table \ref{table:scenenet_sun}. As we can see, our method performs much better than ``No Adapt.", and is only a few points short of the state-of-the-art methods which use source data for adaptation. 

From the tables we can also see that, our method outperforms other source-free adaptation methods, even with just using a batch size of $1$, whereas some baselines use batch size $>1$ to train. Our method is also comparable to many recent methods, which use source data to adapt. Most methods in literature perform an alignment either in the input or output space between the source and target domain, which not only requires the source data, but also additional models (discriminators), which are trained simultaneously. Compared to that, our framework is quite light-weight as it just involves pseudo-label training along with certain constraints. 

{\flushleft \bf{Ablation study of transforms:}} In this experiment, we analyse the effect of transforms we use in our framework and the effect of learning from collage images. We use two spatial transforms, Mirror and Rotate and two augment transforms, Gaussian filter and Cutout. The results are presented in Table \ref{table:transform_ablation}. We evaluate the model by adding only one transform at a time. As can be seen, the transforms individually improve the performance beyond just pseudo-labeling. 
Moreover, learning from collage images also improves the performance beyond pseudo-label learning from single images. The maximum gain is achieved when we use all of the above together.
We also see similar trends for other datasets as well, and present them in appendix (Table~\ref{table:transform_ablation_supp}). It is interesting to note that augment transforms beyond a certain parameter limit do not work as well, which is intuitive, as the content of the image may change beyond what is necessary for fine segmentation. For example, in Gaussian filtering, using a filter size more than $20$ does not help with an image of size $512 \times 1024$.

{\flushleft \bf{Ablation study of the framework:}} In this experiment, we break down every part of the framework and evaluate their performance. We present the results in Table \ref{table:combined_ablation} for all the datasets. When we first update the normalization parameters, the performance improves by $1-4\%$. Then using the updated network for pseudo-label training using the target images offers a further $4-5\%$ improvement. Now, imposing the transform constraints using the hard and soft constraint losses as in Eqn. \ref{eqn:constrain_loss} along with the collage images further improves the performance by about $3\%$ compared to just pseudo-labeling. 

{\flushleft \bf{Ablation of design choices:}} We perform ablation of various choices involved in designing our proposed algorithm. The first one is using category-wise thresholds for pseudo-labeling, compared to uniform thresholding for all labels. 
To compare, we execute our framework with an uniform threshold of $0.9$ for all the labels (following \cite{Li_CVPR_2019}), as mentioned in the sixth row in Table~\ref{table:combined_ablation}. We observe that using label-wise thresholding (``Ours" in Table~\ref{table:combined_ablation}) performs better by $1.5-4\%$ than uniform thresholding.

\begin{table}[!t]
			\small
			\caption{ Ablation of loss functions and design choices.
    	    }
			\label{table:combined_ablation}
			\centering
			\renewcommand{\arraystretch}{1.2}
	        \setlength{\tabcolsep}{7pt}
	        \resizebox{0.48\textwidth}{!}{
			\begin{tabular}{lccc}
		    \toprule
		    \multirow{2}{*}{Methods} & GTA5 & SYNTHIA  & SceneNet\\
		    & $\rightarrow$ Cityscapes & $\rightarrow$ Cityscapes & $\rightarrow$ SUN\\
		    \midrule
		    No Adapt. & 37.5 & 32.1 & 26.5 \\
		    Update Norm & 41.4 & 35.3 & 27.4 \\
		    Pseudo-Label (PL) & 45.6 & 40.2 & 32.1 \\
		    PL + soft constraints & 48.2 & 43.1 & 35.0 \\
		    PL + hard constraints & 47.9 & 43.0 & 34.8 \\
		    \midrule
            Uniform thresholds & 47.3 & 41.6 & 31.9\\
		    Constraints using $\boldsymbol{\textrm{M}}_s$ & 47.3 & 42.2 & 33.2\\
		    Finetuning $\boldsymbol{\textrm{M}}_s$ & 48.6 & 43.3 & 28.9\\
		    PL + augment & 43.1 & 39.1 &  33.8 \\
		    \midrule
		    Ours & \textbf{48.8} & \textbf{43.7} & \textbf{35.6}\\
		    \bottomrule
	    \end{tabular}
	    }
\end{table}

In the next ablation, we study the effect of using the norm updated source model $\boldsymbol{\textrm{M}}_s$ instead of target model $\boldsymbol{\textrm{M}}_t$ in current iteration, for the constraints of Eqn. \ref{eq:augment_transform}, \ref{eq:spatial_transform}. If we apply the transform constraints using $\boldsymbol{\textrm{M}}_s$ rather than $\boldsymbol{\textrm{M}}_t$, then we limit its ability to improve beyond the fixed pseudo-labels, albeit with transforms applied on them. In other words, when using $\boldsymbol{\mathrm{M}}_t$ for the constraints, the optimization process is allowed to learn the target model and self-improve in a way such that the constraints are satisfied on the final target model. This helps to improve the performance, as is evident by comparing the seventh row in Table \ref{table:combined_ablation} with the last row.

Next, we investigate whether to fine-tune the source model for adaptation or train a new target model from scratch. As can be observed from the eighth row of Table \ref{table:combined_ablation}, fine-tuning performs worse than our method where we train a new target model from scratch. Note that the learning rate and the number of training epochs are kept the same for both cases. This can be attributed to the source model being already in a local optima in the loss landscape, thus may be less influenced by the pseudo-label losses and constraints.

Finally, we use the transforms in the standard augmentation setting (but stronger than normal augments), i.e., augment the images, pseudo-label them and learn the target model using them. We use equal portion of original and augmented images, as in our method. The results are presented as ``PL+augment” in Table \ref{table:combined_ablation}, which shows that our consistency based approach performs much better. This is because the pseudo-labels are much better on the original images rather than on the augmented ones, and it is better to transform the pseudo-labeled image, rather than pseudo-labeling the transformed image. 

\begin{table}[!t]
			\small
			\caption{Results for test-time adaptation with a single iteration of optimization at test time.
    	    }
			\label{table:tta_results}
			\centering
			\renewcommand{\arraystretch}{1}
	        \setlength{\tabcolsep}{4pt}
			\begin{tabular}{lccc}
		    \toprule
		    \multirow{2}{*}{Methods} & GTA5 & SYNTHIA  & SceneNet\\
		    & $\rightarrow$ Cityscapes & $\rightarrow$ Cityscapes & $\rightarrow$ SUN\\
		    \midrule
		    No Adapt. & 37.5 & 32.1 & 26.5 \\
		    \midrule
		    Entropy \cite{wang2021tent} & 38.4 & 32.6 & 27.3 \\
		    Likelihood \cite{mummadi2021testtime} & 38.3 & 32.5 & 27.2 \\
            Ours & \textbf{39.3} & \textbf{33.2} & \textbf{27.6}\\
		    \bottomrule
	    \end{tabular}
\end{table}


{\flushleft \bf{Test Time Adaptation:}} 
To further validate the effectiveness of our consistency based self-training approach, we use it for test-time adaptation, which is recently introduce in~\cite{Sun_ICML_2019}. Here, we do not even have any unlabeled target data to adapt. Instead, we use our loss to individually adapt to each test image, resetting the model back to the source model after predicting. For TTA, we do not use the collage transformations, as we adapt to a single image at a time. We show results on the episodic fully test time adaptation setting proposed by \cite{wang2021tent} where the network cannot be trained again with any additional source or target data, other than the given test image. We compare our loss function with those from recent works \cite{mummadi2021testtime, wang2021tent} and observe that our method performs better due to its inherent ability to exploit the regularization provided by semantic segmentation as a task (Table ~\ref{table:tta_results}). All results share the same backbone and hyper-parameters as described in implementation details. For~\cite{mummadi2021testtime}, we report the best result from SLR/HLR loss variants.

\begin{figure}[t]
    \centering
    \includegraphics[scale=0.38]{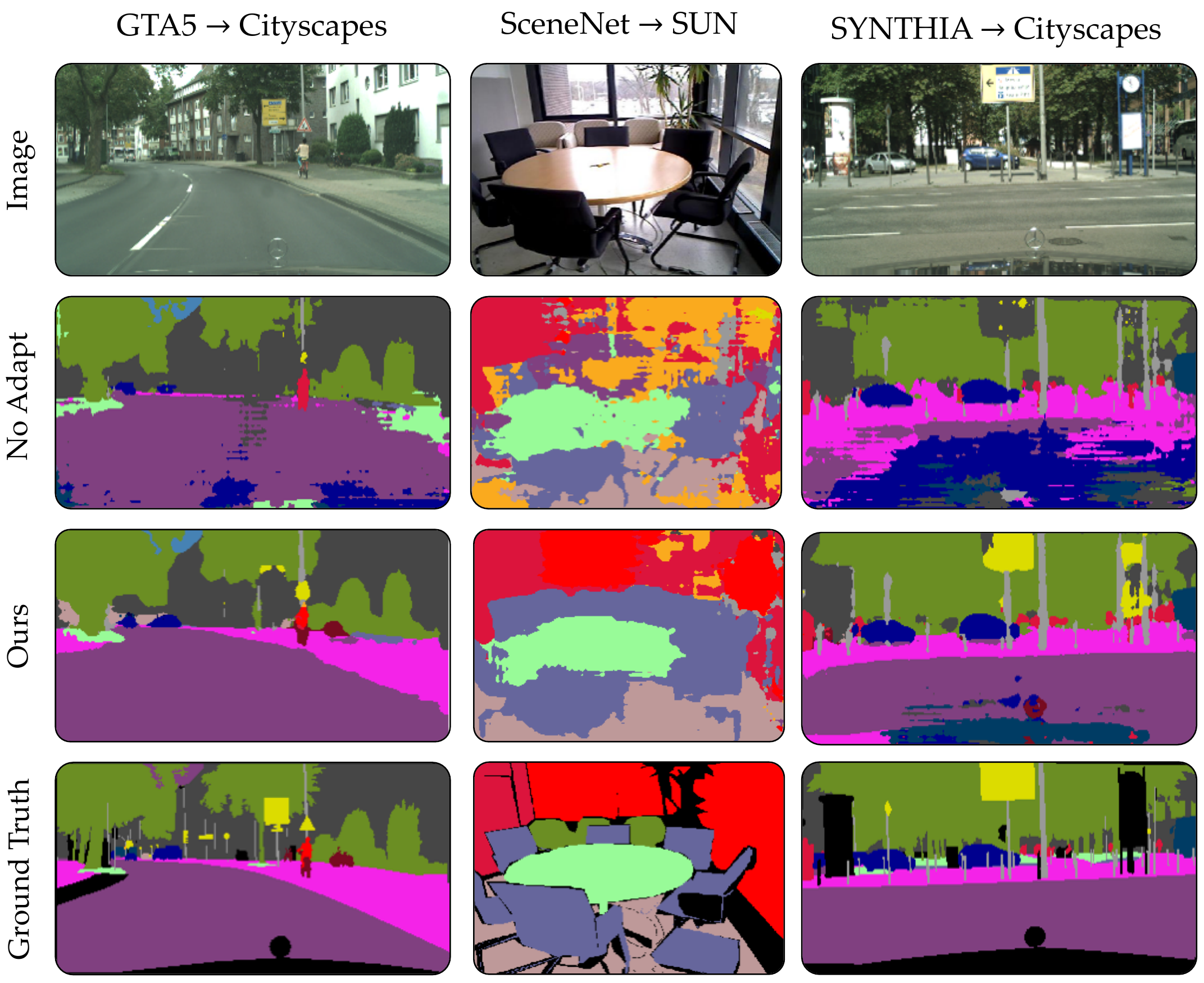}
    \caption{Examples from indoor and outdoor scenarios for visual comparison. The source$\rightarrow$target datasets are mentioned at the top of the columns. ``No Adapt" refers to using the source model directly for prediction, without any adaptation.}
    \label{fig:vis}
\end{figure}

{\flushleft \bf{Qualitative Analysis:}}
We present a few images and their segmentation maps for visual analysis in Figure~\ref{fig:vis}. We compare our method with ``No Adapt", i.e., directly applying the source model. As we can see in the left image, our method is able to discover one of the signs and the sidewalk, which is not segmented in ``No Adapt". For the indoor case, our method correctly segmented the glass regions as window (light red), which is segmented as picture (yellow) in ``No Adapt". In the last column, the signs are correctly discovered by our algorithm, and the road is properly segmented, while it is labeled as car (blue) in ``No Adapt". Overall, the predictions made by our method are smooth with much less noise, which can potentially be attributed to the consistency constraints. We present more examples in appendix (Figure ~\ref{fig:vis_supp}).

\section{Conclusion}

We propose a framework for adapting semantic segmentation models from source to target without access to the labeled source data. We present a self-training framework by enforcing consistencies with certain transforms to efficiently extract information from the source model. With only a few existing works in this challenging setting, our approach compares favorably against strong baselines. 
Empirical evaluation shows that our method performs comparable to many state-of-the-art methods that use source data to adapt.
We further show the usefulness of our approach for fully test-time adaptation in which adaptation is done individually for each test image. 
Future works can explore extracting information such as shape priors, from the source and infusing them into the target model, for better adaptation. 
{\small
\bibliographystyle{ieee_fullname}
\bibliography{bib}
}

\begin{appendices}
\section{Datasets}
\label{app:datasets}
{\flushleft \bf{GTA5 $\rightarrow$ Cityscapes:}} In this setting, we consider GTA5 \cite{Richter_ECCV_2016} as the source and Cityscapes \cite{cityscapes} as the target dataset. The source images are at $760 \times 1280$ resolution, while the target images are used at $512 \times 1024$. The datasets have $19$ categories. The source dataset has $24966$ training images and the target has $2975$ training images on which we perform adaptation, and $500$ validation images. We report the performance on these $500$ validation images as is the usual practice in the literature.

{\flushleft \bf{SYNTHIA $\rightarrow$ Cityscapes:}} In this setting, we consider SYNTHIA \cite{Ros_CVPR_2016} as the souce and Cityscapes \cite{cityscapes} as the target dataset. SYNTHIA contains $9400$ training images. However, unlike GTA5, due to lack of proper annotations for a few categories in SYNTHIA, we remove them from evaluation and report the results for $16$ categories, following the literature. 

{\flushleft \bf{SceneNet $\rightarrow$ SUN:}} Both of the above two settings are for outdoor scenes, and in this setting we consider indoor scenes with SceneNet \cite{mccormac2016scenenet} as the source and SUN \cite{song2015sun} as the target. The SceneNet dataset has around $5$ million simulated images. However, a lot of the images are rather simple with only a few categories in them. Thus, to train the source model, we only choose the top $50,000$ images having the highest number of categories. The SUN dataset contains $5285$ training images on which we perform adaptation and $5050$ test images which are used for evaluation. Both of these datasets contain $13$ categories. Specifically, we use the label transformation available with the SceneNet dataset \footnote{https://github.com/ankurhanda/sunrgbd-meta-data} to map the labels in the SUN dataset such that it matches with the label space of SceneNet.

\section{Category-wise Segmentation Results}
\begin{table*} [t]
	\caption{Results of adapting GTA5 to Cityscapes.
	The top row group are methods which use source data during adaptation, while the bottom row group do not use any source data to adapt. Note that the middle row group use DeepLabv3+ResNet-50 as their backbone while all other methods use DeepLabv2+ResNet-101.
	}
	\label{table:gta5_cityscapes}
	\scriptsize
	\centering
	\renewcommand{\arraystretch}{1.7}
	\resizebox{\textwidth}{!}{
	\begin{tabular}{llcccccccccccccccccccc}
		\toprule
		
		
		Source & Method & \rotatebox{70}{road} & \rotatebox{70}{sidewalk} & \rotatebox{70}{building} & \rotatebox{70}{wall} & \rotatebox{70}{fence} & \rotatebox{70}{pole} & \rotatebox{70}{light} & \rotatebox{70}{sign} & \rotatebox{70}{veg} & \rotatebox{70}{terrain} & \rotatebox{70}{sky} & \rotatebox{70}{person} & \rotatebox{70}{rider} & \rotatebox{70}{car} & \rotatebox{70}{truck} & \rotatebox{70}{bus} & \rotatebox{70}{train} & \rotatebox{70}{mbike} & \rotatebox{70}{bike} & mIoU\\
		
		\midrule
		
		\multirow{7}{*}{{Yes}} & AdaptOutput~\cite{tsai2018learning} & 86.5 & 25.9 & 79.8 & 22.1 & 20.0 & 23.6 & 33.1 & 21.8 & 81.8 & 25.9 & 75.9 & 57.3 & 26.2 & 76.3 & 29.8 & 32.1 & 7.2 & 29.5 & 32.5 & 41.4 \\
		
		& AdvEnt~\cite{Vu_CVPR_2019} & 89.4 & 33.1 & 81.0 & 26.6 & 26.8 & 27.2 & 33.5 & 24.7 & 83.9 & 36.7 & 78.8 & 58.7 & 30.5 & 84.8 & 38.5 & 44.5 & 1.7 & 31.6 & 32.4 & 45.5 \\
		
		& SSF-DAN~\cite{Du_ICCV19} &  90.3 & 38.9 & 81.7 & 24.8 & 22.9 & 30.5 & 37.0 & 21.2 & 84.8 & 38.8 & 76.9 & 58.8 & 30.7 & 85.7 & 30.6 & 38.1 & 5.9 & 28.3 & 36.9 & 45.4 \\
		
		& BDL~\cite{Li_CVPR_2019} & 91.0 & 44.7 & 84.2 & 34.6 & 27.6 & 30.2 & 36.0 & 36.0 & 85.0 & 43.6 & 83.0 & 58.6 & 31.6 & 83.3 & 35.3 & 49.7 & 3.3 & 28.8 & 35.6 & 48.5 \\

		& CAG \cite{zhang2019category} & 90.4 & 51.6 & 83.8 & 34.2 & 27.8 & 38.4 & 25.3 & 48.4 & 85.4 & 38.2 & 78.1 & 58.6 & 34.6 & 84.7 & 21.9 & 42.7 & 41.1 & 29.3 & 37.2 & 50.2 \\
		
		& WeakDA \cite{paul2020domain} & 91.6 & 47.4 & 84.0 & 30.4 & 28.3 & 31.4 & 37.4 & 35.4 & 83.9 & 38.3 & 83.9 & 61.2 & 28.2 & 83.7 & 28.8 & 41.3 & 8.8 & 24.7 & 46.4 & 48.2 \\

        & Stuff \cite{wang2020differential} & 90.6 & 44.7 & 84.8 & 34.3 & 28.7 & 31.6 & 35.0 & 37.6 & 84.7 & 43.3 & 85.3 & 57.0 & 31.5 & 83.8 & 42.6 & 48.5 & 1.9 & 30.4 & 39.0 & 49.2 \\
		
		& FDA \cite{yang2020fda} & 92.5 & 53.3 & 82.3 & 26.5 & 27.6 & 36.4 & 40.5 & 38.8 & 82.2 & 39.8 & 78.0 & 62.6 & 34.4 & 84.9 & 34.1 & 53.1 & 16.8 & 27.7 & 46.4 & 50.4 \\
		& SAC \cite{araslanov2021self} & 90.4 & 53.9 & 86.6 & 42.4 & 27.3 & 45.1 & 48.5 & 42.7 & 87.4 & 40.1 & 86.1 & 67.5 & 29.7 & 88.5 & 49.1 & 54.6 & 9.8 & 26.6 & 45.3 & 53.8 \\
		\midrule
		No & SFDA \cite{Liu_CVPR_2021} & 84.2 & 39.2 & 82.7 & 27.5 & 22.1 & 25.9 & 31.1 & 21.9 & 82.4 & 30.5 & 85.3 & 58.7 & 22.1 & 80.0 & 33.1 & 31.5 & 3.6 & 27.8 & 30.6 & 43.2 \\
		\midrule
        \multirow{3}{*}{No}
        & No Adapt. & 79.7 & 21.8 & 66.8 & 19.3 & 20.6 & 22.8 & 28.9 & 12.9 & 76.3 & 19.5 & 60.9 & 56.2 & 26.6 & 77.8 & 33.3 & 27.7 & 3.9 & 25.0 & 32.5 & 37.5 \\
        & URMA \cite{S_2021_CVPR} & 92.3 & 55.2 & 81.6 & 30.8 & 18.8 & 37.1 & 17.7 & 12.1 & 84.2 & 35.9 & 83.8 & 57.7 & 24.1 & 81.7 & 27.5 & 44.3 & 6.9 & 24.1 & 40.4 & 45.1 \\
        & LD\cite{you2021domain} & 91.6 & 53.2 & 80.6 & 36.6 & 14.2 & 26.4 & 31.6 & 22.7 & 83.1 & 42.1 & 79.3 & 57.3 & 26.6 & 82.1 & 41.0 & 50.1 & 0.3 & 25.9 & 19.5 & 45.5 \\
		& Ours & 89.2 & 37.3 & 82.4 & 29.0 & 23.5 & 31.8 & 34.6 & 28.7 & 84.8 & 45.5 & 80.2 & 62.6 & 32.6 & 86.1 & 45.6 & 43.8 & 0.0 & 34.6 & 54.4 & \textbf{48.8} \\
		
		\bottomrule
	\end{tabular}
	}
\end{table*}

\begin{table*} [t]
	\caption{
		Results of adapting SYNTHIA to Cityscapes.
		The top group are methods which use source data during adaptation, while the bottom row do not use any source data to adapt. mIoU and mIoU$^\ast$ are averaged over 16 and 13 categories. Note that the middle row group use DeepLabv3+ResNet-50 as their backbone while all other methods use DeepLabv2+ResNet-101.
	}
	\label{table:synthia_cityscapes}
	\scriptsize
	\centering
	\renewcommand{\arraystretch}{1.75}
	\resizebox{\textwidth}{!}{
	\begin{tabular}{llcccccccccccccccccc}
		\toprule
		
		
		Source & Method & \rotatebox{70}{road} & \rotatebox{70}{sidewalk} & \rotatebox{70}{building} & \rotatebox{70}{wall} & \rotatebox{70}{fence} & \rotatebox{70}{pole} & \rotatebox{70}{light} & \rotatebox{70}{sign} & \rotatebox{70}{veg} & \rotatebox{70}{sky} & \rotatebox{70}{person} & \rotatebox{70}{rider} & \rotatebox{70}{car} & \rotatebox{70}{bus} & \rotatebox{70}{mbike} & \rotatebox{70}{bike} & mIoU & mIoU$^\ast$ \\
		
		\midrule
		
		\multirow{5}{*}{Yes} & AdaptOutput~\cite{tsai2018learning} & 79.2 & 37.2 & 78.8 & 10.5 & 0.3 & 25.1 & 9.9 & 10.5 & 78.2 & 80.5 & 53.5 & 19.6 & 67.0 & 29.5 & 21.6 & 31.3 & 39.5 & 45.9 \\
		
		& AdvEnt~\cite{Vu_CVPR_2019} & 85.6 & 42.2 & 79.7 & 8.7 & 0.4 & 25.9 & 5.4 & 8.1 & 80.4 & 84.1 & 57.9 & 23.8 & 73.3 & 36.4 & 14.2 & 33.0 & 41.2 & 48.0 \\
		
		& SSF-DAN~\cite{Du_ICCV19} &  84.6 & 41.7 & 80.8 & - & - & - & 11.5 & 14.7 & 80.8 & 85.3 & 57.5 & 21.6 & 82.0 & 36.0 & 19.3 & 34.5 & - & 50.0 \\
		

        & CAG \cite{zhang2019category} & 84.7 & 40.8 & 81.7 & 7.8 & 0.0 & 35.1 & 13.3 & 22.7 & 84.5 & 77.6 & 64.2 & 27.8 & 80.9 & 19.7 & 22.7 & 48.3 & 44.5 & 51.4 \\
		
		& WeakDA \cite{paul2020domain} & 92.0 & 53.5 & 80.9 & 11.4 & 0.4 & 21.8 & 3.8 & 6.0 & 81.6 & 84.4 & 60.8 & 24.4 & 80.5 & 39.0 & 26.0 & 41.7 & 44.3 & 51.9 \\

        & Stuff \cite{wang2020differential} & 83.0 & 44.0 & 80.3 & - & - & - & 17.1 & 15.8 & 80.5 & 81.8 & 59.9 & 33.1 & 70.2 & 37.3 & 28.5 & 45.8 & - & 52.1  \\

		& FDA \cite{yang2020fda} & 79.3 & 35.0 & 73.2 & - & - & - & 19.9 & 24.0 & 61.7 & 82.6 & 61.4 & 31.1 & 83.9 & 40.8 & 38.4 & 51.1 & - & 52.5 \\
		& SAC\cite{araslanov2021self} & 89.3 & 47.2 & 85.5 & 26.5 & 1.3 & 43.0 & 45.5 & 32.0 & 87.1 & 89.3 & 63.6 & 25.4 & 86.9 & 35.6 & 30.4 & 53.0 & 52.6  & 59.3\\
		\midrule
		No
		& SFDA \cite{Liu_CVPR_2021} & 81.9 & 44.9 & 81.7 & 4.0 & 0.5 & 26.2 & 3.3 & 10.7 & 86.3 & 89.4 & 37.9 & 13.4 & 80.6 & 25.6 & 9.6 & 31.3 & 39.20 & 45.9 \\
        \midrule
        \multirow{3}{*}{No}
        & No Adapt. & 37.6 & 18.7 & 73.8 & 9.95 & 0.1 & 26.4 & 8.9 & 13.9 & 74.7 & 80.4 & 52.4 & 16.1 & 39.2 & 21.9 & 13.2 & 25.8 & 32.1 & 36.7  \\
        & URMA \cite{S_2021_CVPR} & 59.3 & 24.6 & 77.0 & 14.0 & 1.8 & 31.5 & 18.3 & 32.0 & 83.1 & 80.4 & 46.3 & 17.8 & 76.7 & 17.0 & 18.5 & 34.6 & 39.6 & 45.0 \\
		
		& LD\cite{you2021domain} & 77.1 & 33.4 & 79.4 & 5.8 & 0.5 & 23.7 & 5.2 & 13.0 & 81.8 & 78.3 & 56.1 & 21.6 & 80.3 & 49.6 & 28.0 & 48.1 & 42.6 & 50.1 \\
        & Ours & 74.3 & 33.7 & 78.9 & 14.6 & 0.7 & 31.5 & 21.3 & 28.8 & 80.2 & 81.6 & 50.7 & 24.5 & 78.3 & 11.6 & 34.4 & 53.7 & \textbf{43.7} & \textbf{50.2} \\
		
		\bottomrule
	\end{tabular}
	}
\end{table*}

In this section, we present the category-wise segmentation results for the two dataset settings, GTA5$\rightarrow$Cityscapes and SYNTHIA$\rightarrow$Cityscapes in Table \ref{table:gta5_cityscapes} and \ref{table:synthia_cityscapes} respectively.

\begin{figure}[h]
    \centering
    \includegraphics[scale=0.5]{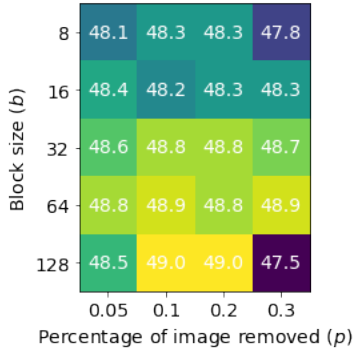}
    \caption{Ablation study of the cutout transformation.}
    \label{fig:cutout_ablation}
\end{figure}

\section{Hyper-parameter analysis of cutout}
In this section, we analyse the effect of the hyper-parameters involved in the cutout transformation. Recall from the paper, in this transformation, we choose two parameters: size $b$ of the square block, and the percentage $p$ of the image to be removed. Given these two parameters, we choose the number of blocks to be removed as $k=\frac{pHW}{b^2}$. We execute our framework for various values of $b$ and $p$, while keeping the rest of the framework same, and present the performance obtained in Figure \ref{fig:cutout_ablation}. By using this transform, we want the network to learn rich context information, while performing inpainting in the output space. As can be seen, with lower block size and higher percentage of image removed, the performance degrades. This is because by doing so the image becomes noisy, rather than structured removal, i.e., removed portions become well distributed throughout the image, which makes it harder to learn context information. However, with higher block size, and moderate percentage of image removed, the performance increases. We choose $b=64, p=0.2$, in our experiments on outdoor images, where the resolution of the image is high, and we choose $b=32, p=0.1$, for indoor experiments, with lower image sizes.

\begin{figure}[h]
    \centering
    \includegraphics[scale=0.5]{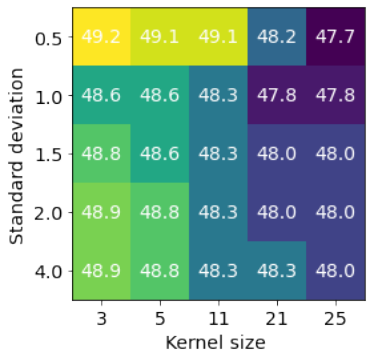}
    \caption{Ablation study of the Gaussian filtering transformation.}
    \label{fig:gauss_ablation}
\end{figure}

\begin{figure*}[!ht]
    \centering
    \includegraphics[scale=0.38]{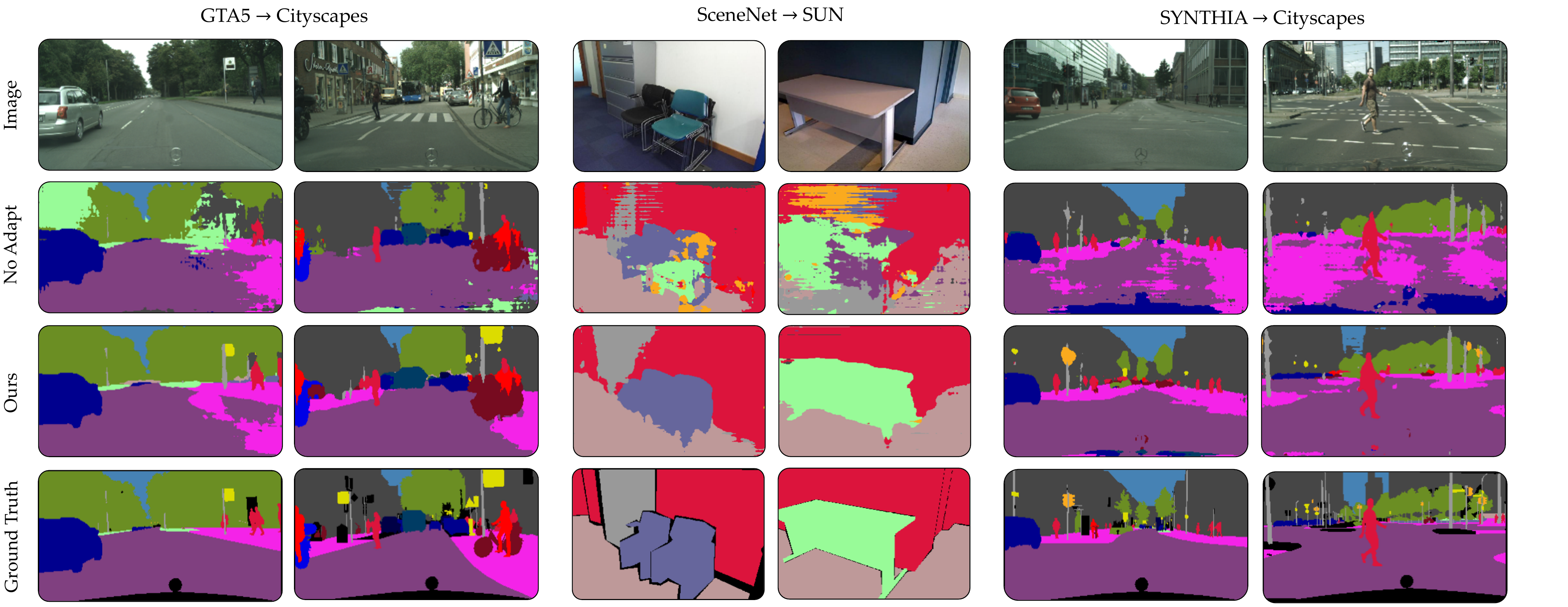}
    \caption{Qualitative visualization. The first two columns and the last two columns show results on outdoor scenes, while the middle two columns show results on indoor scenes.}
    \label{fig:vis_supp}
\end{figure*}

\section{Hyper-parameter analysis of Gaussian filter}

In this section, we analyse the effect of two hyperparameters in the Gaussian filter transformation. The two hyper-parameters are the kernel size and the standard deviation of the filter. Instead of keeping a single standard deviation for the filter, we choose it randomly between [0.1, $\sigma_{max}$], where $\sigma_{max}$ is the hyper-parameter to choose. The image becomes more blurry with higher kernel size and lower standard deviation. As the image becomes more blurry, it becomes difficult for the network to figure out the content, and thus we see a degradation in performance in the top right corner in Figure \ref{fig:gauss_ablation}. Note that the standard deviation mentioned in the figure signify $\sigma_{max}$. We use a kernel size of $5$ and $\sigma_{max}=2$.

\begin{table*}[h]
			\small
			\caption{Ablation of the transformations. 
    	    }
			\label{table:transform_ablation_supp}
			\centering
			\renewcommand{\arraystretch}{1.3}
	        \setlength{\tabcolsep}{7pt}
			\begin{tabular}{ccccccc}
		    \toprule
		    \multirow{1}{*}{Collage} & \multirow{1}{*}{Mirror} & \multirow{1}{*}{Rotate} & \multirow{1}{*}{Gaussian} & \multirow{1}{*}{Cutout} & SYNTHIA & SceneNet \\
		    & & & & & $\rightarrow$ Cityscapes & $\rightarrow$ SUN \\
		    \midrule
		    \checkmark & & & & & 40.4 & 32.2 \\
		    & \checkmark & & & & 41.9 & 34.6 \\
		    & & \checkmark & & & 42.3 & 34.8 \\
		    & & & \checkmark & & 42.5 & 34.5 \\
            & & & & \checkmark & 41.9 & 33.1 \\
            \checkmark & \checkmark & \checkmark & \checkmark & \checkmark &  \textbf{43.7} & \textbf{35.6}\\
		    \bottomrule
	    \end{tabular}
\end{table*}

\section{Ablation of the transformations}

Here, we perform ablation of the four transformations used in our framework - mirror, rotate, gaussian filter and cutout. In the paper, we presented results on GTA5 $\rightarrow$ Cityscapes (Table~\ref{table:transform_ablation}), and here we present on the other two settings, i.e., SYNTHIA $\rightarrow$ Cityscapes and SceneNet $\rightarrow$ SUN. As can be seen in Table~\ref{table:transform_ablation_supp}, each of the transformations have their own contribution towards the final performance, however, an ensemble of all the transformation performs the best.

\section{Additional test time adaptation results}

In this section, we present additional results for the fully test time adaptation setting, where we optimize the network for 5 iterations using the given loss functions per single test image. Note that although not designed specifically for test time adaptation, our consistency based approach is able to perform good on this task as well. As can be seen in Table \ref{table:tta_results}, we perform better or within a close margin to the loss functions proposed by previous works \cite{wang2021tent, mummadi2021testtime} which were completely dedicated towards solving the test time adaptation problem.
\begin{table}[!htbp]
			\small
			\caption{Results for test-time adaptation with a single or five iterations of optimization at test time.
    	    }
			\label{table:tta_results}
			\centering
			\renewcommand{\arraystretch}{1}
	        \setlength{\tabcolsep}{4pt}
			\begin{tabular}{lccc}
		    \toprule
		    \multirow{2}{*}{Methods} & GTA5 & SYNTHIA  & SceneNet\\
		    & $\rightarrow$ Cityscapes & $\rightarrow$ Cityscapes & $\rightarrow$ SUN\\
		    \midrule
		    No Adapt. & 37.6 & 32.1 & 26.5 \\
		    \midrule
		    \multicolumn{4}{c}{1 Iteration} \\
		    \midrule
		    Entropy \cite{wang2021tent} & 38.4 & 32.6 & 27.3 \\
		    Likelihood \cite{mummadi2021testtime} & 38.3 & 32.5 & 27.2 \\
            Ours & \textbf{39.3} & \textbf{33.2} & \textbf{27.6}\\
            \midrule
            \multicolumn{4}{c}{5 Iterations} \\
            \midrule
		    Entropy \cite{wang2021tent} & 40.4 & 34.0 & \textbf{28.8} \\
		    Likelihood \cite{mummadi2021testtime}  & 39.4 & 33.7 & 27.9 \\
            Ours & \textbf{42.2} & \textbf{36.0} & 28.4\\
		    \bottomrule
	    \end{tabular}
\end{table}


\section{Qualitative Analysis}
In Figure \ref{fig:vis_supp} we present a visual comparison of our method with directly applying the source model on the target images, shown as ``No Adapt". As can be seen in the first column, our method is able to properly label the signs, even though the source model does not have them, as shown in ``No Adapt". Similar discussions can be extended to results in the second column. In the third and fourth columns, the source model is very confused with the shadows on the chairs and tables, and assigns them multiple labels. However, our method is able to predict the accurate labels with proper segmentation. In the last column, the network mistakes most portion of the road as sidewalk, a prior that pedestrians can be more often seen on sidewalks than roads. However, our method is able segment the road and sidewalk properly and is also able to segment a few of the road signs in the background.

\end{appendices}

\end{document}